\pdfoutput=1

\documentclass[conference]{IEEEtran}
\ifCLASSINFOpdf
\else
\fi

\pdfoutput=1

\usepackage[left=1in, right=1in, top=1in, bottom=1in]{geometry}
\usepackage{graphicx}
\usepackage{amsmath,amsbsy,amssymb,amsfonts,amsthm}
\usepackage{nicefrac}
\usepackage{mathtools}
\usepackage{color}
\usepackage{xspace} 
\usepackage[numbers]{natbib} 
\usepackage{times}
\usepackage{graphicx,subfigure}
\usepackage{algorithm,algorithmic} 
\usepackage{hyperref}

\newcommand{\noteoff}[1]{}
\newcommand{\grad}{\nabla} 

\newtheorem{theorem}{Theorem}

\newtheorem{assumption}[theorem]{Assumption}

\begin{document}
%
\title{Asynchrony begets Momentum, \\
with an Application to Deep Learning}

\author{
\IEEEauthorblockN{Ioannis Mitliagkas}
\IEEEauthorblockA{Dept. of Computer Science\\
Stanford University\\
Email: imit@stanford.edu}
\and
\IEEEauthorblockN{Ce Zhang}
\IEEEauthorblockA{Dept. of Computer Science\\
ETH, Zurich\\
Email: ce.zhang@inf.ethz.ch} 
\and
\IEEEauthorblockN{Stefan Hadjis, Christopher R\'e }
\IEEEauthorblockA{Dept. of Computer Science\\
Stanford University\\
Email: \{shadjis, chrismre\}@stanford.edu}
}
\maketitle

\begin{abstract}
Asynchronous methods are widely used in deep learning, but have limited theoretical justification when applied to
non-convex problems.
We show that running stochastic
gradient descent (SGD) in an asynchronous manner can be viewed as
adding a momentum-like term to the SGD iteration. Our result does not
assume convexity of the objective function, so it is applicable to
deep learning systems. We observe that a standard queuing model of
asynchrony results in a form of momentum that is commonly used by deep
learning practitioners. This forges a link between queuing theory and
asynchrony in deep learning systems, which could be useful for systems
builders. 
For convolutional neural networks, we experimentally
validate that the degree of asynchrony directly correlates with the
momentum, confirming our main result.
An important implication is that tuning the momentum parameter is important when considering different levels of asynchrony.
We assert that properly tuned momentum reduces the number of steps required for convergence.
Finally, our theory suggests new ways of counteracting the adverse effects of asynchrony: a simple mechanism like using negative algorithmic momentum can improve performance under high asynchrony.
Since asynchronous methods have better
hardware efficiency, this result may shed light on when asynchronous
execution is more efficient for deep learning systems.
\end{abstract}


\IEEEpeerreviewmaketitle

\section{Introduction}
{\color{black}
Stochastic Gradient Descent (SGD) and its variants are the optimization method of choice for many large-scale learning problems including deep learning
\cite{gardner1984learning,amari1993backpropagation,
zhang2004solving,bottou2010large}.
A popular approach to running these systems removes locks and synchronization barriers \cite{recht2011hogwild}. Such methods are called {\em asynchronous-parallel methods} or Hogwild! and are used on many systems by large companies like Microsoft and Google  \cite{dean2012large,chilimbi2014project}.

However, the effectiveness of asynchrony is a bit of a mystery. For convex problems on sparse data, these race conditions do not slow down convergence too much \cite{recht2011hogwild}, and the lack of locking means that each step takes less time. However, sparsity could not be the complete story as many groups have reported that asynchronous-parallel can be faster even for dense data \cite{chilimbi2014project,dean2012large}, in which case available theory \cite{recht2011hogwild,mania2015perturbed} does not apply.
Recent work includes results in the dense case, for general convex problems asymptotically \cite{chaturapruek2015asynchronous}, 
and specifically for matrix completion problems \cite{de2015taming}.

In deep learning there has been a debate about how to scale up training. Many systems have run asynchronously \cite{dean2012large,chilimbi2014project}, but some have proposed that synchronous training may be faster
\cite{abadi2016tensorflow,chen2016revisiting,cui2016geeps}.
As part of our work \cite{hadjis2016omnivore}, we realized that often many systems do not tune the {\em momentum} parameter \cite{polyak1964some}, and this can change the results drastically.
Among deep learning practitioners---and some theoreticians---``momentum'' is a synonym for $0.9$. This is evidenced by the large number of papers and tutorials that prescribe it \cite{krizhevsky2012imagenet,cui2016geeps,caffetutorialsolver,rudertutorial} and by the fact that many high-quality publications \cite{chen2016revisiting,abadi2016tensorflow} do not report the value used, supporting the understanding that $0.9$ has almost reached ``industry standard'' status. That said, there are some papers reporting results after tuning momentum, e.g.,\ \cite{chilimbi2014project}.

Like the step size, the best value for the momentum parameter depends on the objective, the data, and the underlying hardware.
Until now, there was not much reason to think that the optimization and system dynamics interact, although this is folklore among mathematical optimization researchers. In this paper we provide theoretical and experimental evidence that the parameters and asynchrony interact in a precise way. We summarize our contributions:

\begin{itemize}
\item We show that asynchrony introduces momentum to the SGD update, called the {\em implicit momentum}. 
\item We argue that tuning the algorithmic momentum parameter \cite{polyak1964some},
is important when considering different scales of asynchrony.
\item Under a simple model, we describe exactly how implicit and {\em explicit} (algorithmic) momentum interact when both are present. We see numerically that under heavy asynchrony, {\em negative values of explicit momentum} are actually optimal.
\item We verify all of these results with experiments on convolutional neural networks on our prototype system \cite{hadjis2016omnivore}.
\end{itemize}

}

\section{Preliminaries}
\label{sec:model}
Our aim is to minimize an objective $f:\mathbb{R}^d \rightarrow \mathbb{R}$,
of the form
\begin{align}
	\label{eqn:empirical-risk}
	f(w) \triangleq \frac{1}{n} \sum_{i=1}^n f_i(w)
			= \frac{1}{n} \sum_{i=1}^n f(w;z_i).
\end{align}
Typically the component function $f_i(w)$ represents  a loss function evaluated on a specific data point or mini-batch $z_i$. 
SGD considers one term at a time, calculates its gradient and uses it to update vector $w$.
\begin{align}
\label{eqn:sgd}
	w_{t+1} = w_t - \alpha_t \grad_w f(w_t;z_{i_t})
\end{align}
Sequence $(i_t)_t$ describes the order in which the samples are considered and $(\alpha_t)_t$ are the step sizes used.

\subsubsection{Momentum}
Introduced by Polyak \cite{polyak1964some}, the momentum algorithm is ubiquitous in deep learning implementations, as it is known to offer a significant optimization boost \cite{sutskever2013importance}. For some $\mu_L \in [0,1)$ it takes the following form.
\begin{equation}
\label{eqn:momentum}
	 w_{t+1} - w_t
	= \mu_L (w_t - w_{t-1})
	- \alpha_t \grad_w f(w_{t}; z_{i_t})
\end{equation}
In this paper we call $\mu_L$ the {\em explicit
  momentum}; the reason for this name to become clear soon.

\subsubsection{Asynchrony}
A popular way of parallelizing SGD is  the fully asynchronous execution of \eqref{eqn:sgd} by $M$ different workers, also known as \textsc{Hogwild!}   \cite{recht2011hogwild}.
In the simplest topology, all workers share access to a main parameter store; either in main memory or a parameter server.
Worker processes operate on potentially stale values of $w$. Let $v_t$ denote the value read by the worker in charge of update $t$. Then
\begin{equation}
	v_t = w_{t-{\tau}_t},
\end{equation}
for some random delay, ${\tau}_t$, called {\em staleness}. This is referred to as the {\em consistent reads} model \cite{mania2015perturbed} and it makes sense for the dense updates and batch size typically used in CNNs. For the ensuing analysis, we make the assumption that, for every worker and sample $t$, the read delays are independent and follow a distribution denoted by $Q$. Specifically,
\begin{align}
\label{eqn:staleness-distribution}
v_t = w_{t-l} \qquad \textrm{w.p.} \quad q_l,\quad l \in \mathbb{N},
\end{align}
and the update step becomes
\begin{align}
\label{eqn:asgd}
	w_{t+1} = w_t - \alpha_t \grad_w f(v_t;z_{i_t}),
\end{align}
where the gradient is taken with respect to the first argument of $f$.

\begin{figure}[tbp]
\begin{center}
\includegraphics[width=0.50\textwidth]{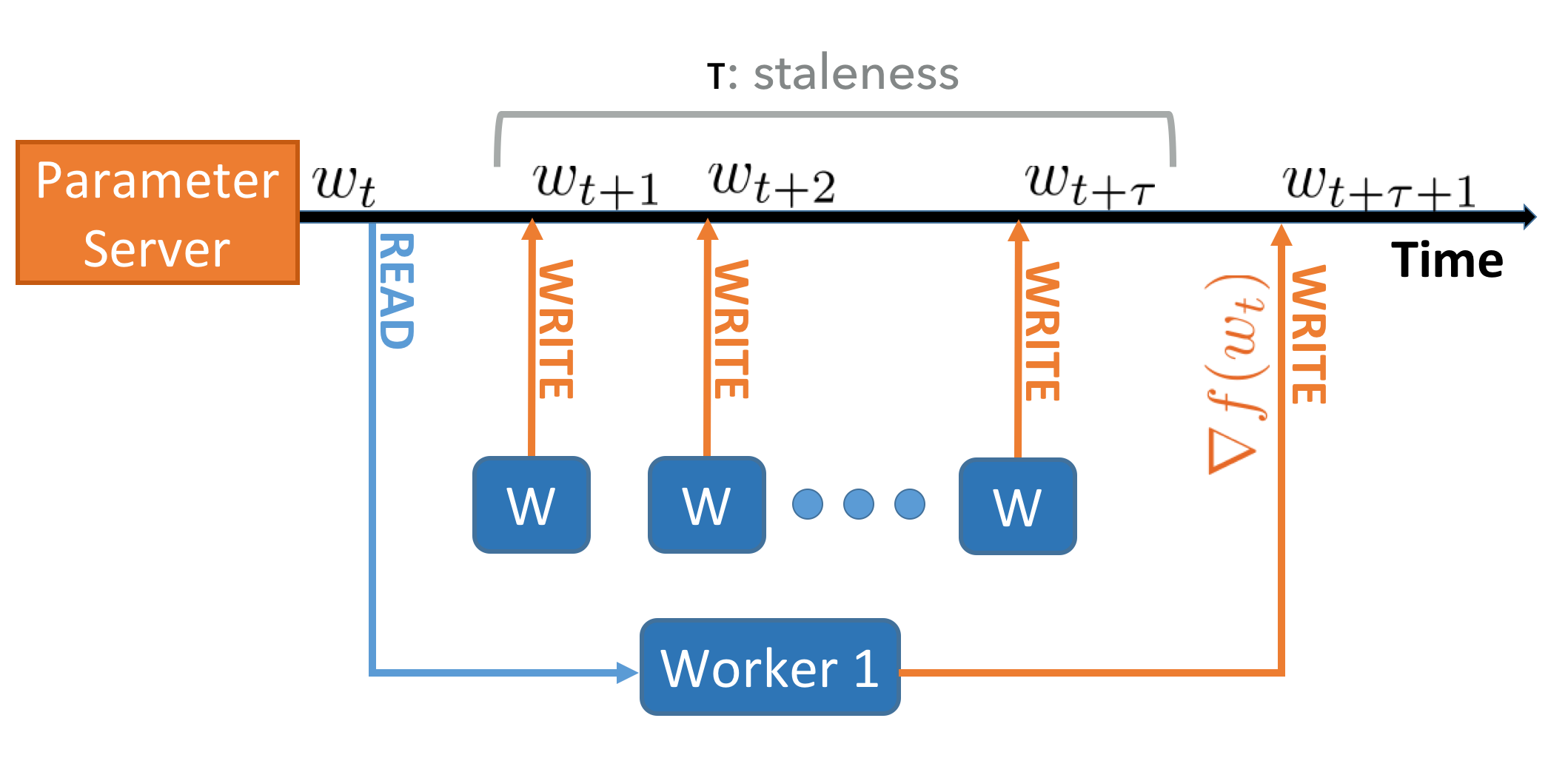}
\end{center}
\vspace{-0.2in}
\caption{Staleness is the number of writes in between a worker's read and write operation. We model it as a random variable $\tau$.}
\label{fig:staleness}
\end{figure}

\section{Asynchrony begets momentum}
The stochastic noise of SGD has been shown to have a stabilizing
effect \cite{hardt2015train}. The effect of asynchrony, on the other
hand, is not well understood and is often assumed to act as a
regularizer. We show that it actually acts as an extra momentum
term. We call this the {\em asynchrony-induced or implicit momentum}
to differentiate from the explicit momentum introduced in
Section~\ref{sec:model}.  Our experimental findings in
Section~\ref{sec:experiments} support this intuition. Specifically,
we see that the optimal value for the explicit momentum drops as we
increase the number of workers. Our understanding is that in those
cases, asynchrony contributes the missing momentum.
We also see that tuning the momentum decreases the number of steps required to reach a target loss.

According to the model described in \eqref{eqn:staleness-distribution}, the value  $v_t$---read and used for the evaluation of step $t$---is a random variable sampled from the model's history, $(w_s)_{s\leq t}$. For example, the expectation of the value read is a convex combination of past values,
$\mathbb{E}[v_t] = \sum_{l=0}^\infty q_l \mathbb{E}[w_{t-l}]$.
This implies the existence of memory in an asynchronous system.
In the following theorem we make this intuition rigorous. 
Some proofs are included in Appendix~\ref{sec:proofsasynchronymomentum};
the rest can be found in the full version of this paper \cite{mitliagkas2016asynchrony}.

\begin{assumption}[Staleness and example selection are independent]
\label{ass:indepstaleness}
We make the following assumption on staleness.
\begin{enumerate}
\item [(A1)]  The staleness process, $(\tau_t)_t$, and the sample 
selection process, $(i_t)_t$, are mutually independent.
\end{enumerate}
\end{assumption}
This assumption is valid on a CNN that performs dense updates, where the randomness in staleness comes from unmodeled implementation and system behavior. 

\begin{theorem}[Memory from asynchrony]
\label{thm:asynchronyismomentum}
Under Assumption~\ref{ass:indepstaleness} and 
for a constant step size $\alpha_t = \alpha$, we get the following momentum-like expression for consecutive updates.
\begin{align*}
	\mathbb{E}[w_{t+1} - w_t]
	=& \mathbb{E}[w_t - w_{t-1}]
		- \alpha q_0 \mathbb{E}\nabla f(w_t)  \\
		&+ \alpha \sum_{l=0}^\infty (q_l - q_{l+1}) \mathbb{E}\nabla f(w_{t-(l+1)})
\end{align*}
\end{theorem}
This theorem suggests that, when staleness has strictly positive variance, the previous step contributes positively to the current step.
Next we show that when staleness is geometrically distributed, 
 we get the familiar form of momentum, discussed in Section~\ref{sec:model}.
\begin{theorem}[Momentum from geometric staleness]
\label{thm:geomstalenessmomentum}
Let the staleness distribution be geometric on $\{0,1,\ldots\}$ with parameter 
$1-\mu_S$, i.e.\ $q_l = (1-\mu_S)\mu_S^l$. The expected update takes the momentum form of \eqref{eqn:momentum}.
\begin{align}
	\mathbb{E}[ w_{t+1} - w_t ] 
	=& \mu_S \mathbb{E}[w_t - w_{t-1}] \notag \\
	&- (1-\mu_S)\alpha \mathbb{E}\grad_w f(w_{t})
\end{align}
\end{theorem}

\subsection{Queuing Model}
\label{sec:queueingmodel}
Here we show that under a simple queuing model, the conditions of Theorem~\ref{thm:geomstalenessmomentum} are satisfied. We denote the time it takes step $t$ to finish  $W_t$, and call $(W_t)_t$ the {\em work process}.

\begin{assumption}[Independent, Exponential Work]
\label{ass:exponentialwork}
We make the following assumptions on the work process.
\begin{enumerate}
\item [(A2)]  $W_{t} \sim \mathrm{Exp}(\lambda)$
\item [(A3)] $(W_{t})_{t}$ are mutually independent.
\end{enumerate}
\end{assumption}

\begin{theorem}
\label{thm:queueing}
Consider $M$ asynchronous workers and 
let $(W_{t})_t$ denote the work process. If the $W_{t}$s are mutually independent and exponentially distributed with parameter $\lambda$, then
\begin{align}
	\mathbb{E}[ w_{t+1} - w_t ] 
	= \left(1 - \frac{1}{M}\right) \mathbb{E}[w_t - w_{t-1}] \notag \\
	\qquad - \frac{1}{M} \alpha \mathbb{E}\grad_w f(w_{t}) 
\end{align}
or equivalently, the asynchrony-induced (implicit) momentum is
$	\mu_S = 1 - \frac{1}{M}$.
\end{theorem}

\begin{figure}[tbp]
\begin{center}
\includegraphics[width=0.5\textwidth]{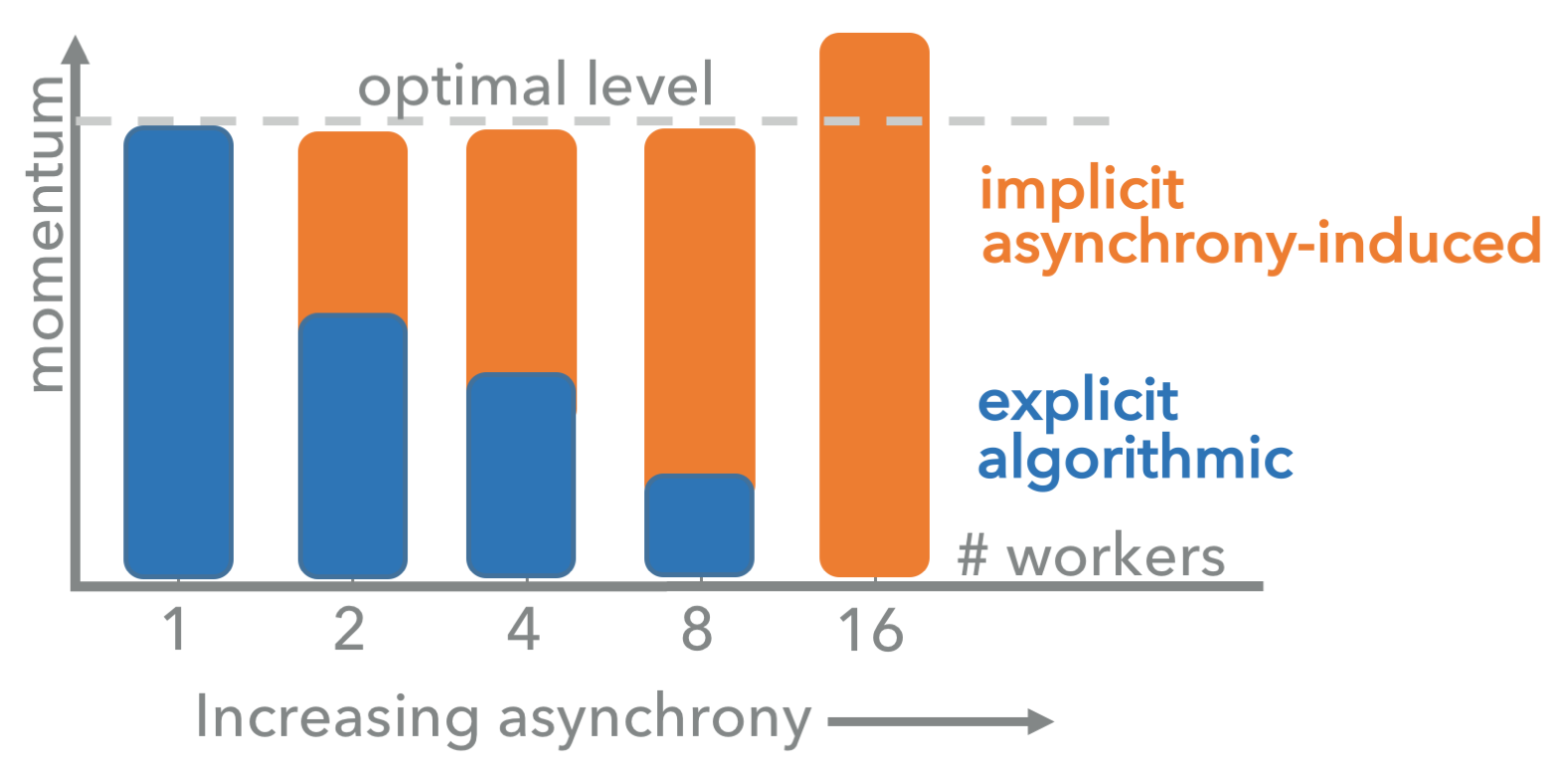}
\end{center}
\vspace{-0.2in}
\caption{Momentum behavior based on queuing model. Total momentum has some optimal value. When asynchrony-induced (implicit) momentum is less than that, we can algorithmically compensate for the
rest. Beyond a certain point, asynchrony causes too much momentum, leading to statistical inefficiency. 
 }
\label{fig:modelmomentum}
\end{figure}

{\color{black}
The theorem suggests that, when training asynchronously, there are two sources of momentum:
\begin{itemize}
\item \textbf{Explicit or algorithmic momentum}: what we introduce algorithmically by tuning the momentum parameter in our optimizer.
\item \textbf{Implicit or asynchrony-induced momentum}: what asynchrony contributes, as per Theorem~\ref{thm:queueing}.
\end{itemize}
For now we can consider that they act additively: the total effective momentum is the sum of explicit and implicit terms. This is a good first-order approximation, but can be improved by carefully modeling their higher-order interactions (cf.\ Section~\ref{sec:counteracting}).
The second theoretical prediction is that more workers introduce more momentum.
Consider the following thought experiment, visualized in Figure~\ref{fig:modelmomentum}.
As we increase the number of asynchronous workers we tune, in each case, for the optimal explicit momentum.
}

This result gives some insight on the limits of asynchrony. 
Consider a case for which the optimal momentum in the
sequential case is $\mu^*$. Theorem~\ref{thm:queueing} tells us that
asynchrony-induced momentum is $\mu_S = 1 - 1/M$, for $M$ asynchronous workers. Therefore there exists an $M_0$ such that $\mu_S>\mu^*$ for all 
$M>M_0$. In other words, too much asynchrony brings about too much momentum to the point of hurting performance.
In Section~\ref{sec:counteracting}, we show that there are ways to counteract these adverse effects of high asynchrony.
In the next section, we validate this section's theoretical findings.

\section{Experimental Validation}
\label{sec:experiments}

We conduct experiments on 9 GPU machines on Amazon EC2 (g2.8xlarge).
Each machine has 4 NVIDIA GRID K520 GPUs.
One of these machines is used as a parameter server, and
the other 8 machines can be organized into 
{$1$, $2$, $4$, $8$} compute groups \cite{DBLP:journals/pvldb/ZhangR14,hadjis2016omnivore}.
Within each group, machines split a mini-batch and combine their updates synchronously. 
Cross-group updates are asynchronous. 
In this paper, we only study how momentum affects the number of iterations to reach a target loss for varying levels of asynchrony.
Hence, groups are equivalent to the workers introduced in Section~\ref{sec:model}.
Our recent paper \cite{hadjis2016omnivore} includes end-to-end performance results; here we present results that pertain to our theory.
We run experiments on two data sets: (i)
CIFAR, and (ii) ImageNet.
For CIFAR, we train the network provided
by Caffe \cite{cifarsolver} and 
for ImageNet, we run CaffeNet \cite{imagenetsolver}.
For ImageNet, we grid search the explicit momentum
in $\{0.0,0.3,0.6,0.9\}$, learning rate
in $\{0.1,0.01,0.001\}$ and set batch size to $256$; for CIFAR we grid search
the explicit momentum in $\{0.0,0.1,...,0.9\}$,
learning rate in $\{0.1,0.01,\ldots,0.00001\}$
and set batch size to $128$.

\begin{figure}[tbp]
\begin{center}
\includegraphics[width=0.4\textwidth]{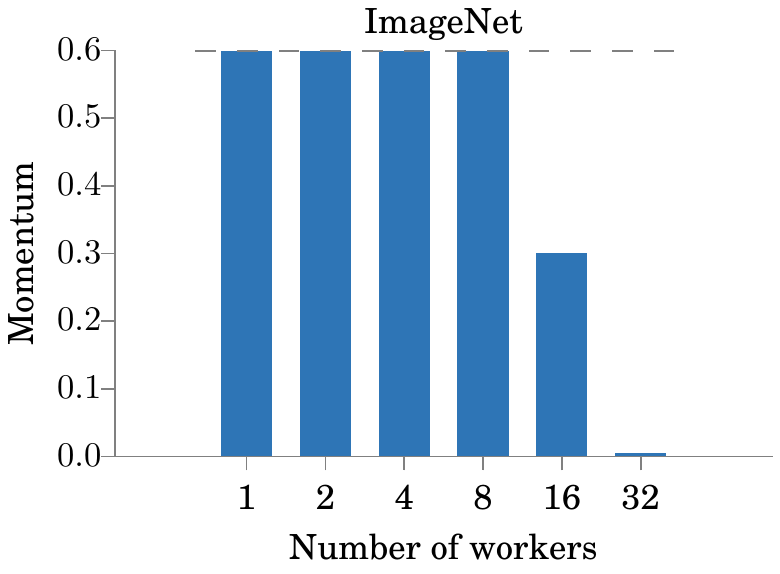}
\vspace{0.1in}
\includegraphics[width=0.4\textwidth]{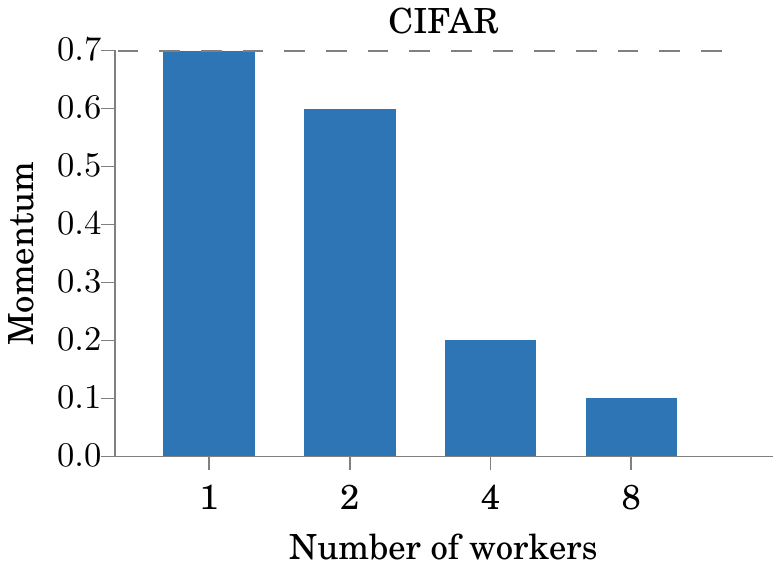}
\end{center}
\vspace{-0.2in}
\caption{Explicit momentum that achieves the best statistical efficiency. (Top)  ImageNet. (Bottom) CIFAR.} 
\label{fig:momentumstaleness}
\end{figure}

Figure~\ref{fig:momentumstaleness} shows the optimal amount of explicit momentum 
(the value that minimizes the number of steps to target loss) as we increase the number of workers. 
The top plot shows the results for ImageNet; the bottom plot shows CIFAR.  We see that our theoretical prediction is
supported by these measurements: the optimal explicit momentum
 decreases when we increase asynchrony.

We have established that the same interactions predicted in our theory manifest in our system, when we tune momentum. 
We now study the performance gains from this tuning process.


\subsection{Measuring Performance}
{\color{black}
The costs and benefits of parallel optimization are best described using the notions of {\em hardware efficiency} and  {\em statistical efficiency}~\cite{DBLP:journals/pvldb/ZhangR14}.
The main performance metric is the wall-clock time to reach a certain training loss or accuracy.
Given a fixed number of machines, we organize them into 
{\em compute groups} \cite{DBLP:journals/pvldb/ZhangR14,hadjis2016omnivore}, a ``hybrid'' between fully synchronous and fully asynchronous configurations.  Within each group, machines combine their updates synchronously. Cross-group updates are asynchronous. We use this architecture in our experiments.
In order to better understand how our design choices affect this metric we decompose it into two factors. The first factor, the number of steps to convergence, is mainly influenced by algorithmic choices and improvements. This factor leads to the notion of {\em statistical efficiency}.
The second factor, the time to finish one step, is mainly influenced by hardware and system optimizations. It leads to the notion of {\em hardware efficiency}.
}


\subsubsection{Hardware Efficiency} 
The obvious benefit of parallelization is ``getting more done'' in the same amount of time. In the case of SGD, we define hardware efficiency to be the 
relative time it takes to complete one step (mini-batch). Specifically, if using one compute group (fully synchronous setting) finishes a batch every $T_1$ seconds, and $m$ groups finish a batch every $T_m$ seconds, the hardware efficiency of using $m$ groups is 
$T_m/T_1$.

\subsubsection{Statistical Efficiency}
On the other hand, some parallelization methods can have a detrimental effect on the quality of the achieved solution. In this case, asynchrony leads to staleness: some gradients are calculated using older models, $w_t$.
Let $I_m$ denote the number of steps required to reach some fixed loss, when using $m$ groups. We define statistical efficiency as $I_m/I_1$. The product of hardware and statistical efficiency is the time it takes $m$ groups to achieve the target accuracy normalized by the corresponding time for a single group; lower values mean faster performance. 

By keeping the total number of workers fixed, we can use these measures of efficiency to study the tradeoffs between different configurations.
Synchronous methods have better statistical efficiency, since all gradients have $0$ staleness; they however suffer from worse hardware efficiency due to waiting at the synchronization barrier. Asynchronous methods provide worse statistical efficiency, but enjoy significant gains in terms of hardware efficiency: there are no stalls.

\section{The importance of tuning}
\label{sec:importance-tuning}
\begin{figure}[htbp]
\begin{center}
\includegraphics[width=0.4\textwidth]{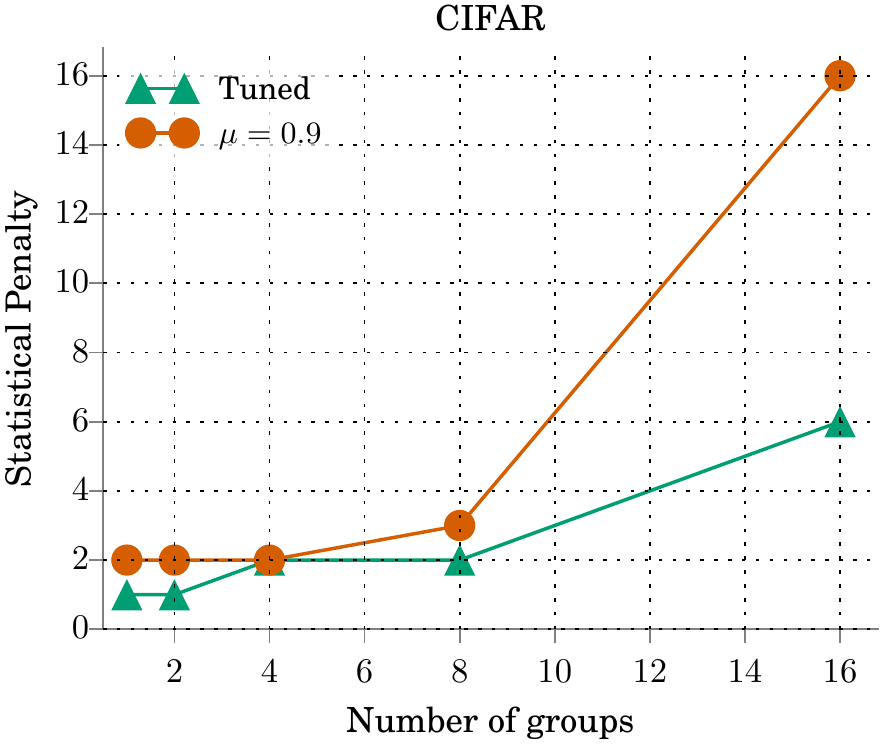}
\end{center}
\vspace{-0.2in}
\caption{The benefits of tuning momentum vs using a commonly prescribed value of $\mu_L=0.9$ on the number of iterations to train the CIFAR dataset. 
 }
\label{fig:tuning-cifar}
\end{figure}

We see that tuning can significantly improve the statistical efficiency of asynchronous training.
In Figure~\ref{fig:tuning-cifar}, we show results on CIFAR.
We conduct experiments on 33 CPU machines
on Amazon EC2 (c4.4xlarge).
One of these machines is used as a parameter server, and
the other machines are organized into compute groups.
We grid search the explicit momentum in $\{-0.9  , -0.675, -0.45 , \ldots,  0.45 ,  0.675,  0.9\}$ and
learning rate in $\{0.1,0.01,0.001,0.0001,0.00001\}$.
We plot the normalized number of iterations for each configuration to reach a target loss, a statistical penalty compared to the best case. We first draw the penalty curve we get by using the standard value of momentum, 0.9, in all configurations. Then we draw the penalty curve we get when we grid-search momentum.
We see that tuning momentum results into an improvement of about $2.5\times$ over the standard value $\mu_L=0.9$.

\begin{figure}[htbp]
\begin{center}
  \includegraphics[width=0.4\textwidth]{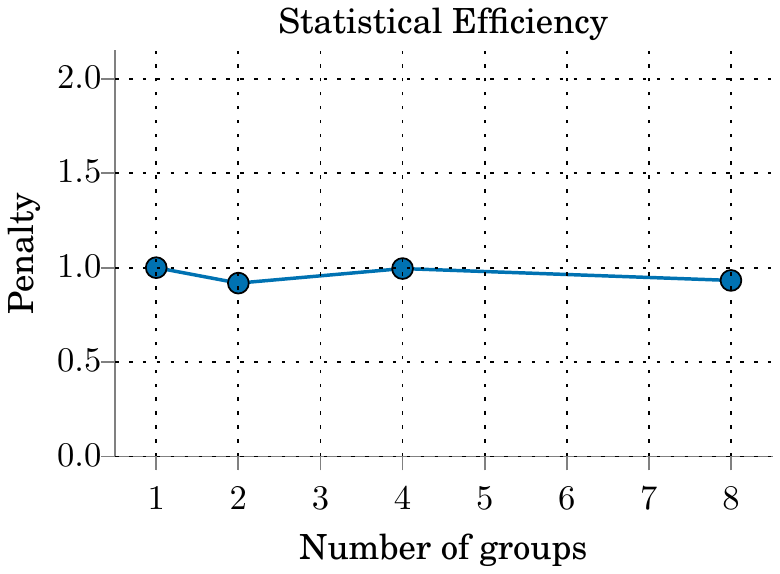}
\end{center}
\vspace{-0.1in}
\caption{Statistical efficiency for ImageNet dataset when tuning \cite{hadjis2016omnivore}. In this case, we pay no penalty for asynchrony.}
\label{fig:heseimagenet}
\end{figure}
{\color{black}
Figure~\ref{fig:heseimagenet} shows the statistical penalty when tuning momentum on ImageNet. The cluster includes $8$ machines, organized into compute groups and the setup is otherwise the same as in Section~\ref{sec:experiments}.
The statistical efficiency result was quite surprising
to us: for ImageNet, even though workers perform dense updates on the model, there is {\em no} statistical efficiency
penalty for up to $8$ groups. This comes in contrast to standard results that say
``asynchrony is fine as long as updates are sparse enough to prevent update collisions.''
 Two comments are in order: (i) this is in an
intermediate stage of execution (not a cold start--which we have observed has
different behavior--but $4000$ iterations into the run), and (ii) there can be a penalty for
some smaller models, or even at larger scales of workers. Our result provides some rough guidance for this behavior. Extensive experiments and detailed setup can be found in our systems paper \cite{hadjis2016omnivore}.
}

\section{Counteracting the effects of asynchrony}
\label{sec:counteracting}

\newcommand{\E}{\mathbb{E}}

In this section we take a closer look at the interaction between asynchrony and momentum. We derive an explicit update rule when both implicit and explicit momentum are present and prescribe a tuning strategy when all parameters are known. Perhaps surprisingly, using negative values of algorithmic momentum can---in some cases---improve the rate of convergence. Then we show experimentally that even when not all parameters are known, tuning via grid search can yield that negative values of algorithmic momentum are the most statistically efficient.

Let us assume the same staleness model of Assumption~\ref{ass:exponentialwork} in the presence of non-zero algorithmic momentum, $\mu_L$.

\begin{theorem}
\label{thm:implicitplusexplicit}
Let the staleness distribution be geometric on $\{0,1,\ldots\}$ with parameter 
$1-\mu_S$, i.e.\ $q_l = (1-\mu_S)\mu_S^l$. For $\alpha' \triangleq (1-\mu_S)\alpha$, the expected update takes the momentum form of \eqref{eqn:momentum}.
\begin{align}
	\mathbb{E}[ w_{t+1}& - w_t ] 
	= (\mu_L+\mu_S) \mathbb{E}[w_t - w_{t-1}] \notag \\
	&-\mu_L\mu_S \mathbb{E}[w_{t-1} - w_{t-2}]
	- \alpha' \mathbb{E}\grad_w f(w_{t})
\end{align}
\end{theorem}


Now we show that convergence rates for quadratic objectives can be easily computed numerically.
\begin{theorem}
\label{thm:quadratic-rates}
Consider a simple quadratic objective 
$f(w)=\frac{1}{2}\|Aw -b\|^2_2$, such that $Aw_*=b$. Let $\lambda_i$ denote the $i$-th eigenvalue of $A^\top A$ and $t_i^*$ denote the root of smallest magnitude for the polynomial
\begin{equation}
	  g_i(t) = \mu_S \mu_L t^3 - \left(	\mu_S + \mu_L + \mu_S \mu_L \right) t^2  + z_i t  -1,
\end{equation}
where 
\begin{equation}
	z_i = 1+\mu_S+\mu_L-\alpha(1-\mu_S)\lambda_i.
\end{equation}
The convergence rate of the expected iterates in the statement of Theorem~\ref{thm:implicitplusexplicit} is given by
\begin{equation}
	\| \E w_t - w_* \|_2 = O(\gamma^t),
	\quad \textrm{ where } \gamma \triangleq \max_i 1/|t_i^*|.
\end{equation}
\end{theorem}

\begin{figure}[tbp]
\begin{center}
\includegraphics[width=0.47\textwidth]{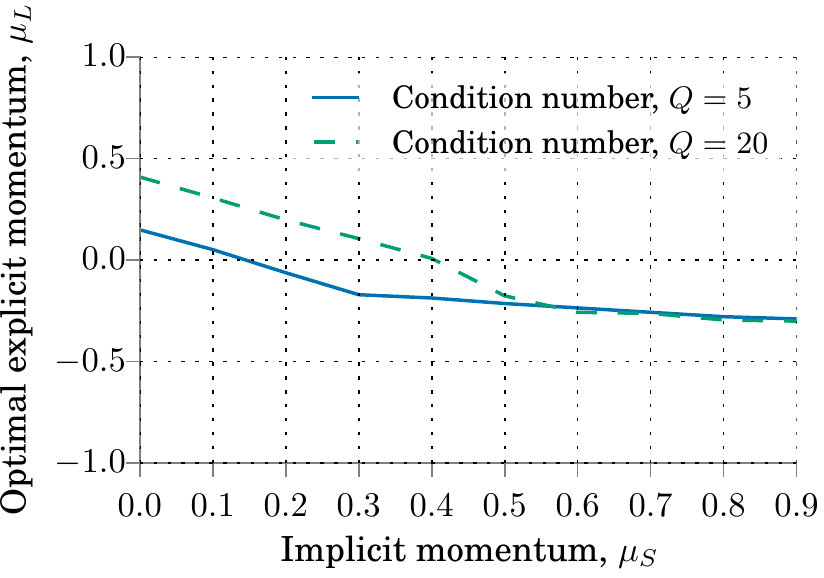}
\end{center}
\vspace{-0.2in}
\caption{Explicit momentum that yields the faster convergence for different values of implicit momentum, caused by staleness. For high implicit momentum, the optimal  explicit momentum is negative.} 
\label{fig:model-optimal-explicit-momentum}
\end{figure}
We can numerically evaluate the rates given in Theorem~\ref{thm:quadratic-rates}, and identify the values of explicit momentum that yield the fastest convergence for a given value of implicit momentum.
Figure~\ref{fig:model-optimal-explicit-momentum} shows the result of a fine grid search over a range of values for explicit momentum $\mu_L$ and step size $\alpha$, for quadratics of condition number $Q=5,20$.
The figure shows that negative values of explicit momentum are optimal when staleness is high.

\begin{figure}[tbp]
\begin{center}
\includegraphics[width=0.47\textwidth]{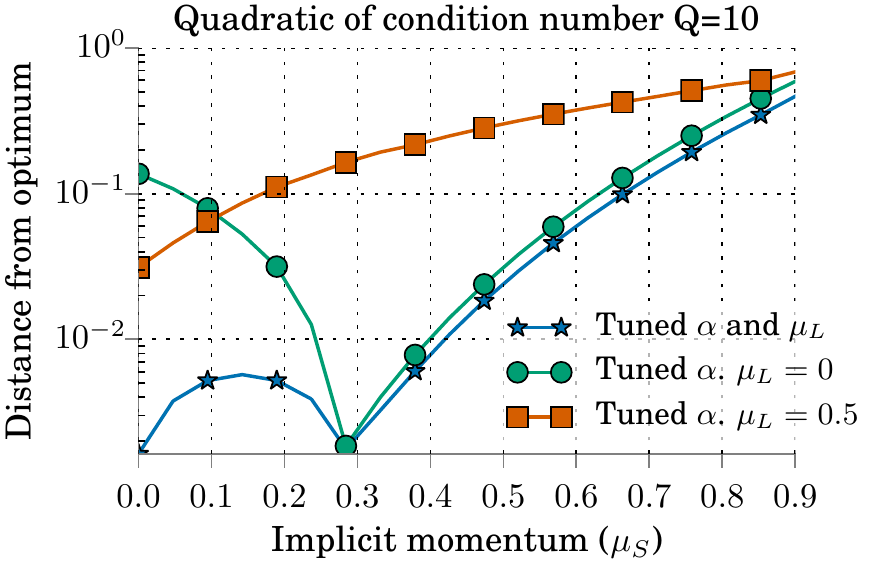}
\end{center}
\vspace{-0.2in}
\caption{Comparing momentum and step size tuning to step size tuning using  rates from Theorem~\ref{thm:quadratic-rates}. For high values of $\mu_S$, momentum tuning is faster than $\mu_L=0.0$ because it selects negative momentum values.}
\label{fig:model-distance}
\end{figure}
In Figure~\ref{fig:model-distance} we use the rates from Theorem~\ref{thm:quadratic-rates} to simulate the result of $10$ steps of the momentum algorithm on a quadratic of condition number $10$. 

We compare three tuning strategies: tuning step size and momentum, versus tuning step size only and fixing momentum to $0$ or $0.5$. We notice that proper tuning makes a difference for both low and high values of implicit momentum. In particular, the use of negative explicit momentum (for $\mu_S>0.3$) results into a speedup of about $1.5x$ compared to using $\mu_L=0.0$.

\begin{figure}[tbp]
\begin{center}
\includegraphics[width=0.47\textwidth]{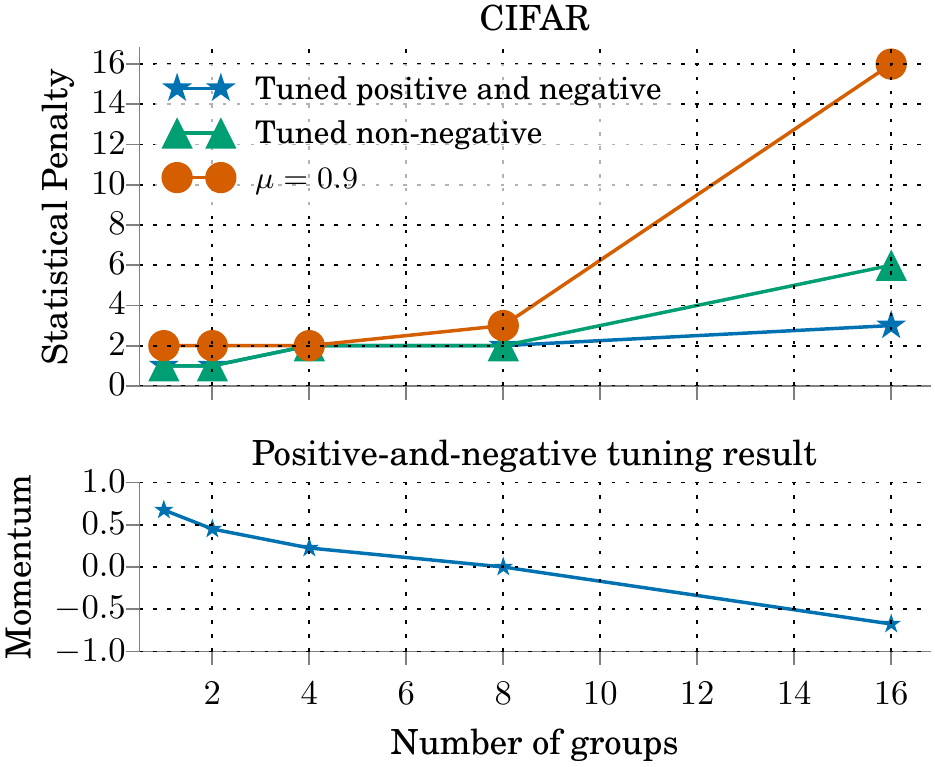}
\end{center}
\vspace{-0.2in}
\caption{Negative momentum tuning experiment on Omnivore and CIFAR. Allowing negative values improves further upon the performance we saw in Figure~\ref{fig:tuning-cifar}.}
\label{fig:negative-momentum-cifar}
\end{figure}
We test this idea on CIFAR on our system \cite{hadjis2016omnivore}, in the setup described in Section~\ref{sec:importance-tuning}.
Figure~\ref{fig:negative-momentum-cifar} shows the statistical penalty (increased number of iterations to the goal) we pay for asynchrony. The top curve is the penalty for using $\mu_L=0.9$. The next one tunes momentum over non-negative values and achieves a speed-up of about $2.5\times$ using 16 groups. When we allow for tuning over negative momentum values, the penalty for $16$ groups improves by another $2\times$.
The bottom plot shows the momentum values selected by the latter tuning process. This provides experimental support to the numerical results Figure~\ref{fig:model-optimal-explicit-momentum} and \ref{fig:model-distance}: negative momentum can reduce the statistical penalty further compared to non-negative tuning.

\section{Discussion and Future Work}

We see that asynchrony-induced  momentum complements algorithmic momentum.
Asynchronous configurations run more efficiently with lower values of momentum.
As a result, any performance evaluation of an asynchronous system needs to take this into account.
A single, globally optimized value of explicit momentum used across configurations of varying asynchrony will yield an inaccurate evaluation: tuning is critical.
Our results suggest simple ways to counteract the adverse statistical effects of asynchrony. A simple technique like negative momentum shows potential of pushing the limits of asynchrony further.
We verified experimentally that the predicted behavior from our simplistic queueing model is present on real systems. Paraphrasing the maxim, any (noise) model is wrong but some are useful.

Our results can be turned into an optimizer that would tune explicit momentum based on current system statistics, like the measured distribution of staleness.
The model presented here was simple and can be extended in many ways,
which we plan to consider in future work.
\begin{itemize}
%

 \item \textbf{Control-theory for momentum compensation} 
 	Ideas like negative momentum seem to work, but we can envision a disciplined way to deal with the adverse effects of asynchrony using control theoretic tools.
  \item \textbf{Topology.} We studied a simple NN topology.
    We can imagine interesting interactions between topology, physical mapping and queueing theory. 
  \item \textbf{Data Sparsity and Irregular Access Patterns.} The work process can depend on the size of the support of the example used. Different applications involve data with different statistics; applications in Natural Language Processing often involve data following heavy-tailed distributions. Also, models with irregular access patterns like LSTMs \cite{hochreiter1997long} may give rise to different staleness distributions.
  \item \textbf{Optimization.} 
  	Different sparsity and staleness distributions naturally lead to momentum different to \eqref{eqn:momentum}. 
Studying the convergence properties on momentum from arbitrary staleness could be of independent theoretical interest.
\end{itemize}

\section*{Acknowledgments} 
 
The authors would like to thank Chris De Sa for the discussion on
asynchrony that was the precursor to the present work
and Dan Iter for his thoughtful feedback.
We would also like to thank
Intel and Toshiba for support along with
the Moore Foundation,
DARPA through MEMEX, SIMPLEX, and XDATA programs,
and the Office of Naval Research, under N000141410102.



\bibliographystyle{IEEEtran}
\bibliography{IEEEabrv,asgd}
%
%
%

\appendix

\section{Proofs}
\label{sec:proofsasynchronymomentum}

\subsection{Proof of Theorem~\ref{thm:asynchronyismomentum}}
\begin{proof}
The statement follows by using \eqref{eqn:asgd} twice,
and subtracting $w_t$ from $w_{t+1}$,
\begin{align*}
	w_{t+1} - w_t =& w_t - w_{t-1}
	- \alpha\Big(
		 \grad_w f(v_t;z_{i_t}) \\
		 &-\grad_w f(v_{t-1};z_{i_{t-1}})
	\Big)
\end{align*}
rearranging terms and taking expectation with respect to the random selection of the $(i_t)_t$'s. This means we are not yet integrating over the randomness in the staleness variables, ${\tau}_t$. Let $\mathcal{T}$ denote the  smallest $\sigma$-algebra under which all the staleness variables are measurable. Then,
using the independence in Assumption~\ref{ass:indepstaleness},
\begin{align*}
	\mathbb{E}[ w_{t+1} - w_t | \mathcal{T}] 
	=& \mathbb{E}[w_t - w_{t-1} | \mathcal{T}]
	- \alpha\Big(
		 \mathbb{E}[ \grad_w f(v_t)  | \mathcal{T}]\\
		 &- \mathbb{E}[ \grad_w f(v_{t-1})  | \mathcal{T}]
		\Big).
\end{align*}
Finally, integrating over all randomness, 
\begin{align*}
	\mathbb{E}[ & w_{t+1} - w_t ] 
	= \mathbb{E}[w_t - w_{t-1}] \\
	&- \alpha\Bigg(
		 \sum_{l=0}^\infty q_l \mathbb{E}\grad_w f(w_{t-l})
		 -\sum_{l=0}^\infty q_l \mathbb{E}\grad_w f(w_{t-l-1})
	\Bigg) \\
	=& \mathbb{E}[w_t - w_{t-1}]
	- \alpha\Bigg(
		q_0 \mathbb{E}\grad_w f(w_{t})\\
		 &+ \sum_{l=1}^\infty q_l \mathbb{E}\grad_w f(w_{t-l})
		 -\sum_{l=0}^\infty q_l \mathbb{E}\grad_w f(w_{t-l-1})
	\Bigg) \\
	=& \mathbb{E}[w_t - w_{t-1}]
	- \alpha q_0 \mathbb{E}\grad_w f(w_{t}) \\
	&- \alpha\Bigg(
		  \sum_{l=0}^\infty q_{l+1} \mathbb{E}\grad_w f(w_{t-l-1}) \\
		 &-\sum_{l=0}^\infty q_l \mathbb{E}\grad_w f(w_{t-l-1})
	\Bigg) \\
	=& \mathbb{E}[w_t - w_{t-1}]
	- \alpha q_0 \mathbb{E}\grad_w f(w_{t}) \\
	&- \alpha\sum_{l=0}^\infty\left(
		   q_{l+1}  - q_l 
	\right)\mathbb{E}\grad_w f(w_{t-l-1}) \\
	=& \mathbb{E}[w_t - w_{t-1}]
	- \alpha q_0 \mathbb{E}\grad_w f(w_{t})\\
	&+ \alpha\sum_{l=0}^\infty\left(
		     q_l - q_{l+1}
	\right)\mathbb{E}\grad_w f(w_{t-l-1})
\end{align*}
\end{proof}

\subsection{Proof of Theorem~\ref{thm:geomstalenessmomentum}}
\begin{proof}
\begin{align*}
	 \mathbb{E}[ w_{t+1} & - w_t ] 
	= \mathbb{E}[w_t - w_{t-1}]
	- \alpha q_0 \mathbb{E}\grad_w f(w_{t}) \\
	&+ \alpha\sum_{l=0}^\infty\left(
		     q_l - q_{l+1}
	\right)\mathbb{E}\grad_w f(w_{t-l-1})\\
	=& \mathbb{E}[w_t - w_{t-1}]
	- \alpha c \mathbb{E}\grad_w f(w_{t}) \\
	&+ \alpha\sum_{l=0}^\infty\left(
		     c \mu^l - c \mu^{l+1}
	\right)\mathbb{E}\grad_w f(w_{t-l-1}) \\
	=& \mathbb{E}[w_t - w_{t-1}]
	- \alpha c \mathbb{E}\grad_w f(w_{t}) \\
	&+  (1 - \mu) \alpha \sum_{l=0}^\infty
		      c \mu^l 
	\mathbb{E}\grad_w f(w_{t-l-1}) \\
	=& \mathbb{E}[w_t - w_{t-1}]
	- \alpha c \mathbb{E}\grad_w f(w_{t}) \\
	&-  (1 - \mu)  \mathbb{E}[w_t - w_{t-1}]\\
	=& \mu \mathbb{E}[w_t - w_{t-1}]
	- \alpha c \mathbb{E}\grad_w f(w_{t}) \\
	=& \mu \mathbb{E}[w_t - w_{t-1}]
	- (1-\mu)\alpha \mathbb{E}\grad_w f(w_{t}) 
\end{align*}
\end{proof}

\subsection{Proof of Theorem~\ref{thm:queueing}}
\begin{proof}

Let $W_{t}$  denote the time the $t$-th iteration takes.
Under Assumption~\ref{ass:exponentialwork}, $W_{t} \sim Exp(\lambda)$. We want the staleness distribution $\tau_t$: the number of writes in the time between the read and write of the reference worker in charge of step $t$. We call this random variable 
$B_t$. These in-between writes are performed by the remaining $M-1$ workers, and 
under Assumption~\ref{ass:exponentialwork}, we get that in $T$ units of time, the number of writes is
\begin{equation}
	B_t(T) \sim \mathrm{Poisson}(\lambda(M-1)T).
\end{equation}
The staleness distribution is the number of writes by other workers in $W_{t}$ units of time.
\begin{equation}
	\tau_t \sim B_t(W_{t})
\end{equation}
It is a simple probability exercise\footnote{For example, \textless$9.7$\textgreater\ in:
\url{http://www.stat.yale.edu/~pollard/Courses/241.fall97/Poisson.Proc.pdf}} to show that $\tau_t$ is geometrically distributed on $\{0,1,\ldots\}$.
\begin{equation}
	{\tau}_t \sim \textrm{Geom}(p),
	\quad p = \frac{\lambda}{\lambda + (M-1)\lambda} = \frac{1}{M}
\end{equation}
where $M$ is the number of workers. Note that $\E \tau_t=M-1$.
Using this with Theorem~\ref{thm:geomstalenessmomentum} we get the statement.
\begin{align*}
	\mathbb{E}[ w_{t+1} - w_t ] 
	=& \left(1 -  \frac{1}{M}\right) \mathbb{E}[w_t - w_{t-1}] \\
	&-  \frac{\alpha}{M} \mathbb{E}\grad_w f(w_{t}) 
\end{align*}
\end{proof}

\onecolumn 

\subsection{Proof of Theorem~\ref{thm:implicitplusexplicit}}

\begin{proof}
We start with the update rule,
\begin{align}
	w_{t+1} = w_t - \alpha \grad f(w_{t-\tau_t}) + \mu_L (w_t - w_{t-1})
\end{align} 
and consider the {\em expected step}, 
\begin{align}
      \label{eqn:mom-stale}
	\E w_{t+1} = \E w_t - \alpha \sum_{l=0}^k q_l \E\grad f(w_{t-l}) + \mu_L \E[w_t - w_{t-1}].
\end{align}
Staleness is geometrically distributed and independent for each write, i.e.\ $q_l = (1-\mu_S)\mu_S^l$ for $l \in \mathbb{N}_0$. We first rearrange the last equation 
\begin{align}
\label{eqn:mom-stale-unwrap}
	\E w_{t+1} - \E w_t - \mu_L \E[w_t - w_{t-1}] = - \alpha \sum_{l=0}^k (1-\mu_S)\mu_S^l \E\grad f(w_{t-l}).
\end{align}
Now starting from \eqref{eqn:mom-stale},
\begin{align*}
	\E w_{t+1}
	=& \E w_t - \alpha \sum_{l=0}^k (1-\mu_S)\mu_S^l \E\grad f(w_{t-l}) + \mu_L \E[w_t - w_{t-1}]\\
	=& \E  w_t - \alpha (1-\mu_S) \E\grad f(w_{t})  - \alpha \sum_{l=1}^k (1-\mu_S)\mu_S^l \E\grad f(w_{t-l}) + \mu_L \E [w_t - w_{t-1}]\\
	=& \E w_t - \alpha (1-\mu_S) \E\grad f(w_{t})  -  \mu_S \alpha \sum_{l=0}^{k-1} (1-\mu_S)\mu_S^{l} \E\grad f(w_{t-1-l}) + \mu_L \E [w_t - w_{t-1}]
\end{align*}
and using \eqref{eqn:mom-stale-unwrap},
\begin{align*}
	\E w_{t+1}
	=&\ \E w_t - \alpha (1-\mu_S) \E\grad f(w_{t}) 
	 +  \mu_S \big(
	 	\E w_t - \E w_{t-1} - \mu_L \E [w_{t-1} - w_{t-2}] 
	 \big)
	 + \mu_L \E [w_t - w_{t-1}]
\end{align*}
and finally,
\begin{align}
	\E w_{t+1} =&\ (1 + \mu_S + \mu_L)\E w_t
		 - \alpha (1-\mu_S) \E\grad f(w_{t}) 
	- \left(
		\mu_S + \mu_L + \mu_S \mu_L
	\right)\E  w_{t-1}
	 +  \mu_S \mu_L \E w_{t-2}
\end{align}
This recurrence holds for $k \geq 2$.
\end{proof}

\subsection{Proof of Theorem~\ref{thm:quadratic-rates}}
\begin{proof}
We will use the polynomial family $q_{k}(z)$ to describe the behavior of $\E[w_k-w_*]$.
From $Aw_*=b$, we get $\nabla f(w) = A^\top A(w-w_*)$.
Then, the statement of Theorem~\ref{thm:implicitplusexplicit}, can be equivalently described by the following polynomial recursion.
\begin{align}
	\label{eqn:mom-stale-poly}
	q_{k+1}(z) =&\ z q_k(z)
	- \left(
		\mu_S + \mu_L + \mu_S \mu_L
	\right) q_{k-1}(z)
	 +  \mu_S \mu_L q_{k-2}(z)
\end{align}
for and $k \geq 2$, 
where $z=(1 + \mu_S + \mu_L)I - \alpha (1-\mu_S) A^TA$.
Note that  $q_0(z)=1$, $q_1(z)=z-\mu_L-\mu_S$ and $q_2(z)=z^2 - (\mu_L+\mu_S)z-(\mu_L+\mu_S)$.
To get the generating function for this recurrence, we multiply by $t^{k+1}$ and sum over $k=2,\ldots\infty$.
\begin{align*}
	\sum_{k=2}^\infty 
		q_{k+1}(z) t^{k+1}
		=&\ 	\sum_{k=2}^\infty z y_k(z) t^{k+1}
	- 	\sum_{k=2}^\infty \left(
		\mu_S + \mu_L + \mu_S \mu_L
	\right) y_{k-1}(z) t^{k+1}
	 +  \mu_S \mu_L 	\sum_{k=2}^\infty w_{t-2}(z) t^{k+1}
\end{align*}
\begin{align*}
	G(z) - q_2(z) - q_1(z) - q_0(z)
		=&\ 	z t (G(z) - q_1(z) - q_0(z))
	- 	\left(
		\mu_S + \mu_L + \mu_S \mu_L
	\right) t^2 (G(z) -  q_0(z))
	 +  \mu_S \mu_L t^2	G(z)
\end{align*}
Rearranging, 
\begin{align*}
	G(z) &\left( 1 -  zt + \left(	\mu_S + \mu_L + \mu_S \mu_L \right) t^2 -  \mu_S \mu_L t^3  \right)	\\
		=&\
	  q_2(z) + q_1(z) + q_0(z)
		-z t (  q_1(z) + q_0(z))
	+ 	\left(
		\mu_S + \mu_L + \mu_S \mu_L
	\right) t^2  q_0(z) \\
		=&\  q_2(z) + q_1(z) + 1
		-z t (  q_1(z) +1)
	+ 	\left(
		\mu_S + \mu_L + \mu_S \mu_L
	\right) t^2 \\
		=&\ 
  z^2 - (\mu_L+\mu_S)z-(\mu_L+\mu_S) + z-\mu_L-\mu_S + 1
		-z t (  z-\mu_L-\mu_S +1)
	+ 	\left(
		\mu_S + \mu_L + \mu_S \mu_L
	\right) t^2 \\
			=&\  
				\left[
		\mu_S + \mu_L + \mu_S \mu_L
	\right] t^2 
	-\left[z  (  z-\mu_L-\mu_S +1)\right]t
	+ \left[
	  z^2 + (1-\mu_L-\mu_S)z +  1-2(\mu_L+\mu_S) 
	  \right]		
\end{align*}
\begin{align*}
	G(z)	
		=&\ 
		-\frac{
				\left[
		\mu_S + \mu_L + \mu_S \mu_L
	\right] t^2 
	-\left[z  (  z-\mu_L-\mu_S +1)\right]t
	+ \left[
	  z^2 + (1-\mu_L-\mu_S)z +  1-2(\mu_L+\mu_S) 
	  \right]	
	  }
	  {
	  \mu_S \mu_L t^3 - \left(	\mu_S + \mu_L + \mu_S \mu_L \right) t^2  + zt  -1
	  }	
\end{align*}
The roots of the denominator (the growth polynomial) of the generating function, dictate the rate of convergence.
In particular, the inverse of the largest root magnitude gives us the desired rate.
Let $A^\top A = Q \Lambda Q^T$ be the eigendecomposition of $A^TA$, with eigenvalues $\lambda_i$. Let $v_t = Q^T w_t$ and note that for every $i$ we get a scalar recurrence for $v_t(i)$.
\begin{align}
	\E v_{t+1}(i) =&\ (1 + \mu_S + \mu_L)\E v_t(i)
		 - \alpha (1-\mu_S) \lambda_i (v_t(i)-v_*(i)) 
	- \left(
		\mu_S + \mu_L + \mu_S \mu_L
	\right)\E  v_{t-1}(i)
	 +  \mu_S \mu_L \E v_{t-2}(i)
\end{align}
According to the analysis above, its growth polynomial is
\begin{equation}
	  g_i(t) = \mu_S \mu_L t^3 - \left(	\mu_S + \mu_L + \mu_S \mu_L \right) t^2  + z_i t  -1,
\end{equation}
where 
\begin{equation}
	z_i = 1+\mu_S+\mu_L-\alpha(1-\mu_S)\lambda_i.
\end{equation}
Now let $t_i^*$ denote the root of smallest magnitude for $g_i(t)$. The rate of convergence along the $i$ eigendirection is $\gamma_i=O(1/|t_i^*|)$. The rate for $w_t$ is dominated by the largest $\gamma_i$, which yields the statement.
\end{proof}

\end{document}